\documentclass[11pt,letterpaper]{article}
\usepackage{cogsys}
\usepackage[T1]{fontenc}
\usepackage{times}
\usepackage[pdftex]{graphicx} 

\usepackage{natbib}
\cogsysheading{11}{2021}{1--**}{9/2021}{11/2021}

\ShortHeadings{Convolutional Cobweb}
              {C.J.\ MacLellan and H.\ Thakur}

\begin{document} 

\title{Convolutional Cobweb: A Model of Incremental\\ 
Learning from 2D Images}
 
\author{Christopher J. MacLellan}{christopher.maclellan@drexel.edu}
\author{Harshil Thakur}{ht377@drexel.edu}
\address{College of Computing and Informatics, Drexel University, 
         Philadelphia, PA 19104 USA}
\vskip 0.2in
 
\begin{abstract}This paper presents a new concept formation approach that supports the ability to incrementally learn and predict labels for visual images. This work integrates the idea of convolutional image processing, from computer vision research, with  a concept formation approach that is based on psychological studies of how humans incrementally form and use concepts. We experimentally evaluate this new approach by applying it to an incremental variation of the MNIST digit recognition task. We compare its performance to Cobweb, a concept formation approach that does not support convolutional processing, as well as two convolutional neural networks that vary in the complexity of their convolutional processing. This work represents a first step towards unifying modern computer vision ideas with classical concept formation research.
\end{abstract}

\section{Introduction}
Recent years have seen an explosion in the availability of image data and, as a result, there has been rapid advances in the development of computer vision approaches.
One kind of approach that has achieved state-of-the-art performance on a large number of computer vision tasks are Convolutional Neural Networks, or CNNs \citep{lecun2015deep}.
These models operate by convolving relatively small (e.g., 3x3) filters over an image to generate transformed feature maps.
Since each filter is applied at every location in the image, the model is able to recognize and respond to patterns that occur both within and across input images. 
Convolutional filters can represent useful transformations, such as edge detection, sharpening, embossing, and blurring.

Although CNNs tend to achieve high performance, they have multiple limitations that are the focus of the current work.
First, although it is possible to train them incrementally, most research efforts  exclusively utilize batch training paradigms.
Second, they require a large amount of training data to achieve high performance (e.g., tens to hundreds of thousands of examples).
If sufficient labeled data is not available, then their performance suffers.
Finally, CNNs have a static structure that is fixed at creation time, only the network weights are updated during training.

To overcome these limitations, we draw on ideas from the concept formation literature \citep{fisher2014concept,gennari1989models}.
Concept formation approaches, such as Cobweb \citep{fisher1987knowledge}, were developed based on psychological studies of how humans incrementally learn concepts.
As a result, they make incremental processing a necessary characteristic.
Further, unlike CNN's which are tuned using stochastic gradient descent---a process that requires many iterative updates---most concept formation approaches use closed-form optimization (e.g., closed-form maximum likelihood estimation), which does not require iterative training to converge.
Lastly, concept formation approaches support the ability to update both their structure and their parameters during online training, whereas CNNs only update their parameters.

Despite these benefits, there do not yet exist many examples of research that apply concept formation approaches to image data, such as the kind of data typically employed by CNNs.
For example, researchers have not typically explored the use of approaches like Cobweb \citep{fisher1987knowledge,mckusick1990cobweb,maclellan2016trestle} or SAGE \citep{mclure2015extending} for learning from pixel-level feature representations.
Additionally, unlike CNNs, concept formation approaches do not make use of convolutional filters to process their inputs.

This work aims to bring together ideas from both of these lines of research.
In particular, we explore the development of a new concept formation approach that we call Convolutional Cobweb.
This approach utilizes convolutional processing to support incremental concept formation over 2D images.
We evaluate our approach on an incremental variant of the MNIST digit recognition task and compare it to three other approaches: Cobweb/3 (a variation of Cobweb that supports continuous features; \citeauthor{mckusick1990cobweb}, \citeyear{mckusick1990cobweb}) as well as two CNNs that vary in terms of the complexity of their convolutional processing.
Based on our empirical study, we make three explicit claims: (1) our new approach's ability to engage in convolutional processing lets it outperform a comparable Cobweb approach that does not support convolutional processing; (2) by utilizing the Cobweb algorithm for categorization and learning, our new approach is able to outperform a simple CNN that utilizes a similar convolutional structure; and (3) our approach outperforms the CNNs when learning incrementally from just a few examples (e.g., less than 100 examples).




\section{The Cobweb Algorithm}

\begin{figure}
    \centering
    \includegraphics[width=0.6\textwidth]{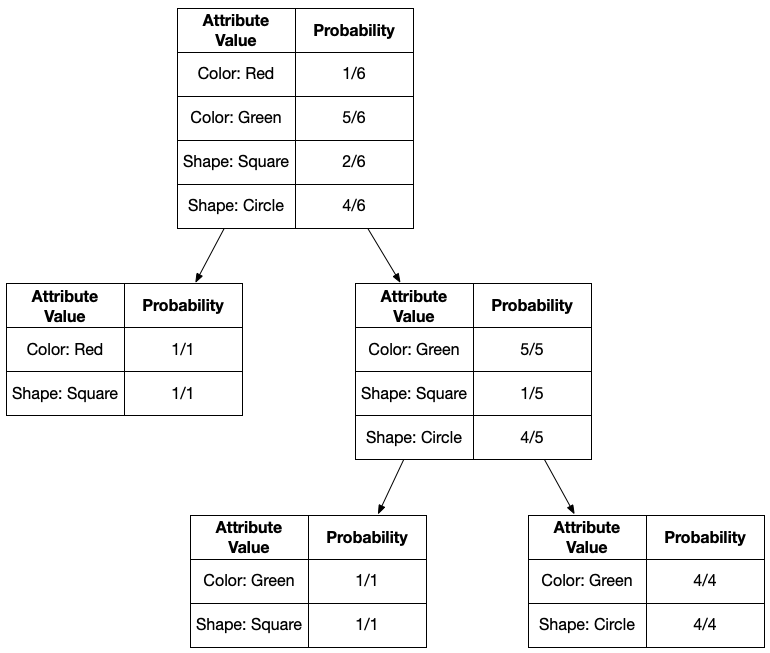}
    \caption{An example of a conceptual hierarchy that Cobweb might learn. Each concept stores a table of attribute-value counts (shown as probabilities) that describe all of the instances stored under the concept.}
    \label{fig:cobweb_example}
\end{figure}

Cobweb \citep{fisher1987knowledge} is an incremental concept formation approach that was developed based on how people incrementally learn concepts.
This approach accepts a series of instances described in an attribute-value representation (e.g., \{color: red, shape: triangle, ...\}).
Given these examples, Cobweb produces a conceptual hierarchy like the one shown in Figure \ref{fig:cobweb_example}.
Each concept in this hierarchy contains a table of attribute-value counts that describe how frequently each attribute-value pair occurs in the instances labeled with the respective concept.

Given a concept hierarchy, Cobweb generates predictions about a new instance by first sorting it down the concept tree. At each point in the sorting process, Cobweb uses a measure called {\it category utility} to determine whether it is better to terminate at the current node or recursively sort the instance into a child concept. The category utility measure is similar to the information gain measure used in decision tree learning \citep{quinlan1986induction}, but optimizes for the ability to predict the values of all attributes rather than a single attribute. This measure was developed based on psychological studies of human concept learning \citep{corter1992explaining}.
Cobweb uses the attribute-value count table from the final concept the instance was sorted into to make predictions about the instance (e.g., to predict the shape of an instance given only its color). 

To update its conceptual hierarchy in light of a new instance, Cobweb uses a procedure similar to the one used at performance time.
Specifically, it sorts the through the instance into its categorization tree using an expanded set of four operations: {\bf add}, which adds the instances to the best child concept (updating the attribute value counts); {\bf merge}, which merges the two child concepts that most closely match the instance; {\bf split}, which splits the concept that most closely matches the instance; and {\bf create}, which creates a new child concept to match the instance.
At each point Cobweb chooses the operation that maximizes the category utility measure.

While Cobweb only supports the ability to process instances that contain nominal attribute values, there have been many extensions that offer support of other kinds of attributes.
For example, Cobweb/3 supports the ability to handle continuous attribute values \citep{mckusick1990cobweb} by approximating them with Gaussian distributions. 
This approach stores the mean and standard deviation of each continuous attribute in each concept description (rather than the attribute-value counts) and uses a modified version of category utility that supports continuous attributes.
Other variations, such as Labyrinth \citep{thompson1991concept} and Trestle \citep{maclellan2016trestle}, support the ability to handle component and relational attributes.

\section{Convolutional Cobweb}

At its core, our new approach is a variation of Cobweb/3 that supports the ability to engage in convolutional processing over 2D image data.
However, we do not simply make use of the convolutional processing approach used by neural networks.
Instead, we have developed a more unified way to integrate convolutional processing in a fashion that is compatible with the spirit of Cobweb.
Prior Cobweb research generally assumes higher-level input representation (e.g., features refer to shapes and colors rather than pixels) and centers the attention on learning concept hierarchies over these higher-level representations.
In contrast, Convolutional Cobweb accepts lower-level pixel features and learns a hierarchy of convolutional filters to represent these inputs as well as a classification hierarchy for making predictions about instances.
In this section, we describe Convolutional Cobweb's representations for example instances as well as learned knowledge. We also describe its performance and learning mechanisms.

\subsection{Instance Representation}

\begin{figure}
    \centering
    \includegraphics[width=\textwidth]{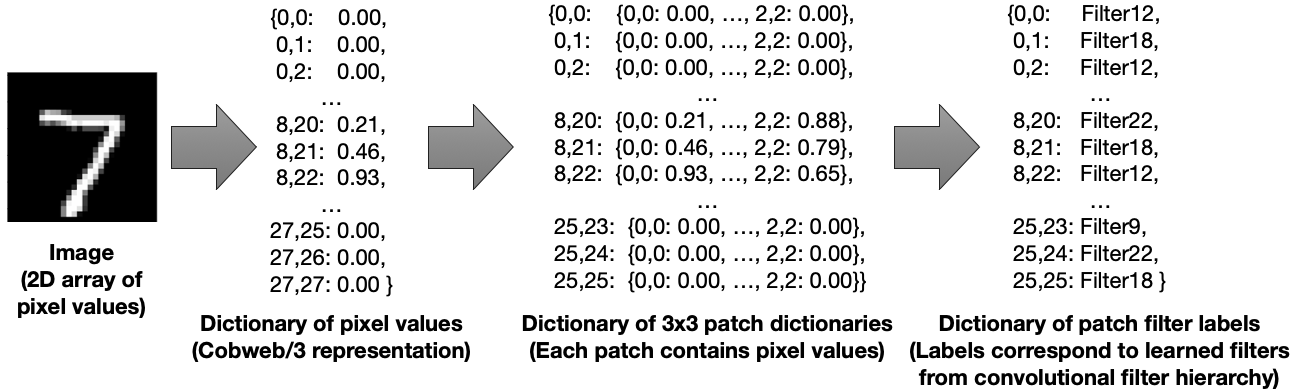}
    \caption{Representations used by Convolutional Cobweb to encode and categorize images.}
    \label{fig:representation}
\end{figure}

Single channel (black and white) image data is generally represented as a 2D array of pixel intensity values.
Convolutional Cobweb translates data input in this format through multiple representations during processing, see Figure \ref{fig:representation}.
First, it translates the 2D array of pixels into a dictionary of attribute-values, where the attribute names represent the indices of the pixels in the original 2D array.
This representation is also the one that we use later in our baseline Cobweb/3 approach.
Next, Convolutional Cobweb converts this representation into a dictionary of all possible patches from the input image that have a user-specified convolutional filter size.
In the current work, we use a filter size of 3x3, so this structure holds all 3x3 patches from the input image.
Each patch is indexed by the location of the upper left pixel in the patch.
Thus, a 28x28 image will generate 26x26 entries, each containing a 3x3 patches.
Each 3x3 patch contains the intensity values for the pixels it covers, indexed by where the pixel occurs in the patch.
Each of these patches is then categorized using a variant of Cobweb that learns a hierarchy of convolutional filters. This categorization process (described in the following performance mechanism subsection) returns a filter label for each patch.
The final representation, which replaces each patch with its respective filter label, is then used for categorization and for making inferences about the image itself (e.g., for predicting the digit label).

\subsection{Knowledge Representation}

\begin{figure}
    \centering
    \includegraphics[width=0.80\textwidth]{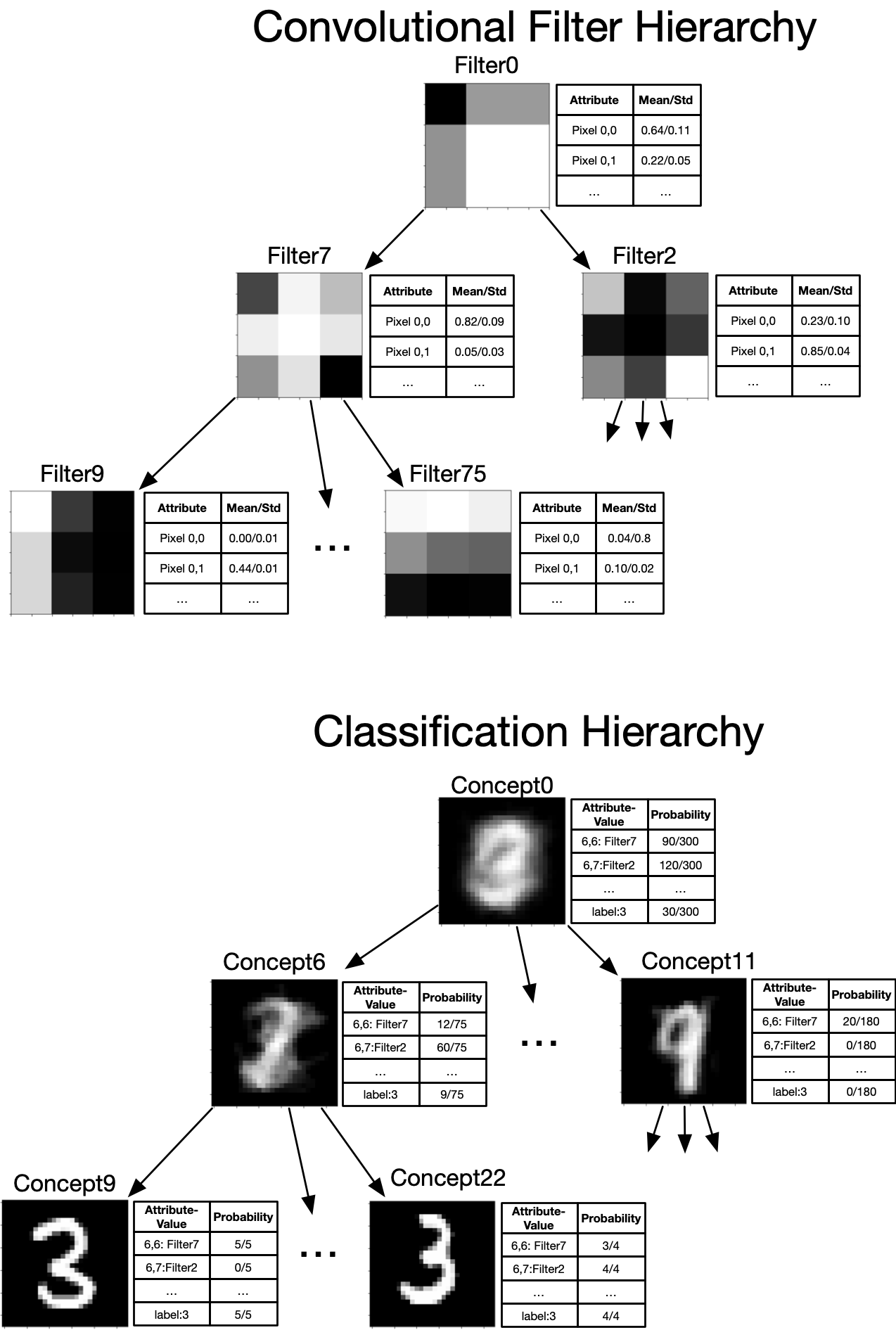}
    \caption{Examples of the two internal hierarchies learned by Convolutional Cobweb. Shown on the top is a hierarchy of 3x3 convolutional filters learned directly from the images. Shown on the bottom is a hierarchy of concepts learned over the convolutional filter representations.}
    \label{fig:convo_cobweb_trees}
\end{figure}

Unlike Cobweb which consists of a single concept hierarchy, Convolutional Cobweb consists of two hierarchies: a convolutional filter hierarchy and a classification hierarchy (see Figure \ref{fig:convo_cobweb_trees}).
The first hierarchy stores a collection of 3x3 convolutional filters learned from the input images.
Similar to Cobweb/3, each node stores the means and standard deviations of the pixel features for all the patches that have been categorized under it.
The second hierarchy stores each image (represented in terms of filters from the convolutional filter hierarchy) and any other features of the image, such as its digit label.
As shown in Figure \ref{fig:convo_cobweb_trees}, each node maintains an table of attribute-value counts\footnote{Although we only show nominal attributes in this example, our implementation can also support continuous attributes using the same approach adopted by Cobweb/3.}.

\subsection{Performance Mechanism}

Leveraging these representations, Convolutional Cobweb uses a variant of Cobweb's performance mechanism to make predictions about each input image.
When it is presented with an input image, such as the image of a 7 shown in Figure \ref{fig:representation}, it first converts this image into a dictionary of 3x3 patch dictionaries (second to last representation from Figure \ref{fig:representation}).
Next, each patch dictionary is sorted through the convolutional filter hierarchy.
At each step in the sorting process, Convolutional Cobweb chooses between stopping at the current filter or sorting recursively into one of the children nodes using the category utility measure.
The instance description is then updated by replacing each 3x3 patch dictionary with the label of the filter it was categorized as (producing the final instance representation from Figure \ref{fig:representation}).

The resulting description is sorted through the classification hierarchy to determine the final concept for the instance.
Convolutional Cobweb sorts the instance through the classification hierarchy using category utility to determine when to stop and when to recursively categorize the instance into a child node.
For each category utility calculation, Convolutional Cobweb dynamically replaces all filter references in the attribute-value count tables with references to filters that are one level below the root of the convolutional filter hierarchy.
For example, references to Filters 9 or 75 from the convolutional filter hierarchy in Figure \ref{fig:convo_cobweb_trees} would be replaced with references to Filter 7 and any associated attribute-value counts would be merged when computing category utility.
This remapping procedure ensures that all concepts use the same level of representation for filters.
Additionally, by using a dynamic remapping approach, Convolutional Cobweb ensures that all filter references are valid, even in situations where the convolutional filter hierarchy is modified (intermediate filters might be created or destroyed) during incremental learning .
Once a final concept has been identified in the classification hierarchy, Convolutional Cobweb uses its attribute-value table to make predictions about the image (e.g., to predict its digit label).


\subsection{Learning Mechanism}

During learning, Convolutional Cobweb goes through a similar process to performance.
It first categorizes each image patch into its convolutional filter hierarchy using the four standard Cobweb operations (add, merge, split, create).
As it incorporates each patch into the hierarchy, it updates the parameters (mean and std of pixels) of relevant filters to reflect the new patch.
After categorizing each patches, it replaces the patch dictionaries with their respective filter labels and adds any additional attribute values of the image (e.g., its digit label) to the instance description.
Convolutional Cobweb then uses the add, merge, split, and create operations to incorporate this updated instance into its classification hierarchy.
To decide which operation is best it uses the modified category utility calculation that dynamically remaps all filter references in the attribute-value count tables to point to the filters that are one level below the root of the convolutional filter hierarchy.

During learning, every unique patch from each image will be represented by a leaf in the convolutional filter hierarchy and every unique image will be represented as a leaf in the classification hierarchy (categorization always ends in the creation of a new leaf or incorporation of the instance into an existing leaf that is identical).
As a result, all references to filters in the classification hierarchy always point to leaves of the convolutional filter hierarchy.
Thus, information about filters is never lost, even when intermediate (non-leaf) filters in the convolutional hierarchy are deleted.
Convolutional Cobweb dynamically remaps all filter references to point to valid intermediate representations.
Figure \ref{fig:convo_cobweb_trees} shows examples of a convolutional filter hierarchy and classification hierarchy that results from applying this approach to a sequence of MNIST images that represent the digits 0-9.
We can see that the convolutional filter hierarchy represents lower-level pixel patterns, whereas the categorization hierarchy uses the filter representation to learn categories of images.

\section{Experimental Evaluation}
Our approach combines ideas from CNNs and Cobweb, so we were interested in how it compares to these baselines.
We conducted experiments to explore whether our approach could outperform Cobweb/3, which uses a similar concept formation approach, but does not support convolutional processing.
We also conducted a set of experiments to evaluate how our approach compares to CNNs, which support convolutional processing, but use gradient descent to update their parameters and do not support the ability to modify their internal structure.

\subsection{Incremental MNIST Digit Recognition Task}
\begin{figure}
    \centering
    \includegraphics[width=\textwidth]{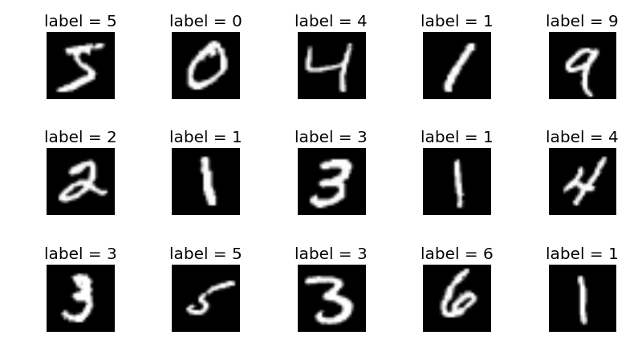}
    \caption{Examples of MNIST images with their labels}
    \label{fig:mnist_examples}
\end{figure}

To evaluate our approach, we chose to test it using images from the Modified National Institute of Standards and Technology (MNIST) data set, which is widely used to test image-based classification approaches \citep{deng2012mnist}.
This data set contains images of hand written digits that are labeled with values ranging from 0 through 9.
Each image contains 28x28 black and white pixel values (784 continuous feature values), which we normalized to have a mean of 0 and a standard deviation of 1.
See Figure \ref{fig:mnist_examples} for a set of example images used in our analysis.

The conventional approach to evaluating classification approaches on MNIST images is to train on a batch of images and then to test a learned model on another batch of held-out cases (e.g., training on 80\% of the images and testing on 20\%).
However, a goal of the current work is to evaluate how well our proposed approach supports incremental prediction and learning.
Thus, we used an incremental prediction paradigm to evaluate our approach and to compare it to baselines.
In this framework, we present each approach with a sequence of images. The first time we present an image from the sequence, we withhold the image label and predict the missing label. Once a prediction has been generated, we then provide each system with both the image and its label and we incrementally update the system's knowledge to account for the new training example. This process is repeated for each image in the sequence until all images have been processed. To estimate the average performance of each approach, we applied this evaluation approach 50 times. For each iteration, we used a sequence of 300 images (30 images of each digit) that was randomly sampled from the MNIST data set and randomly shuffled. To reduce performance differences due to differences in training sequences, we trained each system on the same set of 50 sequences.

\subsection{Comparison Approaches}
To evaluate Convolutional Cobweb, we applied it to the incremental MNIST digit recognition task and compared it to three baseline approaches: (1) Cobweb/3, (2) a simple CNN that has a convolutional structure similar to Convolutional Cobweb, and (3) a more complex CNN with multiple convolutional layers and filters.

\subsubsection{Cobweb/3}
Our first baseline approach is Cobweb/3, which accepts a sequence of images that are represented as a dictionaries of attribute-values (see second representation in Figure \ref{fig:representation}).
Cobweb/3 does not use any specialized approaches for image processing, it simply treats the 784 pixels from each image as continuous features. 
For each input image, it attempts to predict the label for the image using the standard Cobweb prediction algorithm.
Specifically, the instance is sorted into the classification tree and the probability table from the final concept the instance is sorted into is used to generate a prediction of the image label.
While generating predictions, the classification tree is not modified.
However, once a prediction has been generated, the image label is added as an attribute value of the instance and inserted into the tree using the standard Cobweb learning algorithm.
This incrementally updates the tree structure and attribute-value counts by adding, merging, splitting, and creating concepts during instance categorization.
Once the example has been incorporated into the tree, the model is presented with the next image in the sequence.

\subsubsection{CNN-Simple}
Our second approach is CNN-Simple, which uses a neural network structure consisting of 1 convolutional layers and two fully connected layers.
The convolutional layer applies a single 3x3 convolutional filter to each single channel (black and white) image using a padding of 0 and a stride of 1.
The output from the convolutional layer is passed through a non-linear ReLU activation function before being flattened into a vector of features.
These features are then passed through two fully connected layers.
The first layer contains 128 hidden nodes and utilizes an ReLU activation function. 
The second layer contains 10 output nodes and utilizes a Softmax activation function, which translate each output into a probability value.
These 10 output probabilities correspond to the probability that each input image represents one of the ten digits (0-9).

As a preliminary test of this approach, we trained it using a standard CNN training paradigm.
We divided the 70,000 MNIST training images into into 56,000 training images and 14,000 validation images (an 80-20 split).
We used stratification while creating this split to ensure the distribution of each digit was similar in both the training and validation sets.
We then trained the model using stochastic gradient decent with the Adam optimizer \citep{kingma2014adam}.
We trained the model on the entire training set 150 times (i.e., 150 epochs) using a batch size of 1024 images and a cross-entropy loss function.
Once training completed, we then applied the trained model to generate predictions for each of the 14,000 validation images.
We compared these prediction to the ground truth labels to compute the error rate of the model.
This batch training approach shows that CNN-Simple achieves an an error rate of only 2.48\% (97.52\% accurate) on the validation set.

To support the incremental MNIST digit recognition task, we utilized a non-standard approach to train the neural network. 
For each image in the training sequence, we applied the neural network to predict the label of the image.
Once a prediction had been made, we added the image and its associated label to a buffer.
This buffer holds 16 images and their associated labels using a first-in-first-out queue.
After inserting the image into the buffer, we apply the neural network to predict the labels for all 16 images, compute the cross-entropy loss over these predictions, and then apply a single stochastic gradient decent update (using the Adam optimizer) to adjust the network weights.
Thus, when processing an incremental training sequence of 300 images, we apply a total of 300 weight updates to the network.

We designed CNN-Simple to be similar to our Convolutional Cobweb approach, which uses a single convolutional processing step followed by a classification step.
Similarly, CNN-Simple uses a single 3x3 convolutional layer followed by two fully connected layers for classification. 
While it is standard practice within the computer vision community to use use multiple convolutional layers that each have multiple convolutional filters \citep{lecun2015deep}, we believe that this provides a reasonable baseline for comparison to our new Cobweb variant.

\subsubsection{CNN}
Our last approach uses a CNN structure that is more complex than CNN-Simple and is more representative of the kinds of CNNs that typically appear in practice. 
This network utilizes 2 convolutional layers, a max pool layer, and two fully connected layers.
The first convolutional layer maps a single (black and white) input channel to 32 output channels.
Each output channel is constructed by passing 1 of 32 3x3 convolutional filters over the input image with a stride of 1 and padding of 0 (each filter generates its own output channel). 
Each output channel is then passed through an ReLU activation function.
The second convolutional layer maps from 32 channels (output from the first convolutional layer) to 64 output channels.
To handle the multi-channel input, this approach has 32 3x3 convolutional filters for each of the 64 output channels---for a total of 2,048 3x3 convolutional filters.
Each output is generated by applying the output channel's 32 3x3 convolutional filters to their respective 32 input channels and summing the results.
These 64 output channels are then passed through an ReLU activation function and a 2x2 max pool layer, which reduces the height and width of the 64 output channels by half.
The final 64 channels are flattened into a single vector of features and then passed through two fully connected layers that have the same structure as those used by CNN-Simple (128 hidden nodes with an ReLU activation function and 10 output nodes with a Softmax activation function).

Similar to CNN-Simple, we tested this approach using conventional CNN batch training.
Specifically, we divided the MNIST data into training and validation sets using a stratified 80-20 split.
We trained the CNN on the entire training set 150 times (150 epochs) using stochastic gradient decent with the Adam optimizer and then tested it on the validation sets to determine the error rate.
This analysis shows that the more complex CNN is able to achieve an error rate of only 1.26\% (98.74\% accurate) on the validation set.
Thus, when using batch training, this approach achieves better performance on the MNIST data set than the Simple-CNN approach.

To apply this approach to incrementally predict the MNIST image labels, we use the same approach we employed for CNN-Simple.
Specifically, we applied the CNN to predict each image in the sequence, after generating a prediction we added the image and its label to a buffer containing the past 16 images and their associated labels, and then we do a stochastic gradient decent update using the 16 examples in the buffer.
This approach uses a much more complex structure than Convolutional Cobweb or CNN-Simple.
Whereas these other approaches use a single convolutional layer with a single output channel (a single 3x3 filter in the case of CNN-Simple), this approach uses two convolutional layers containing 2,080 3x3 convolutional filters (32 in first layer and 2,048 in the second).
We chose to include this approach in our evaluation because it is representative of typical CNNs.

\begin{figure}
    \centering
    \includegraphics[width=\textwidth]{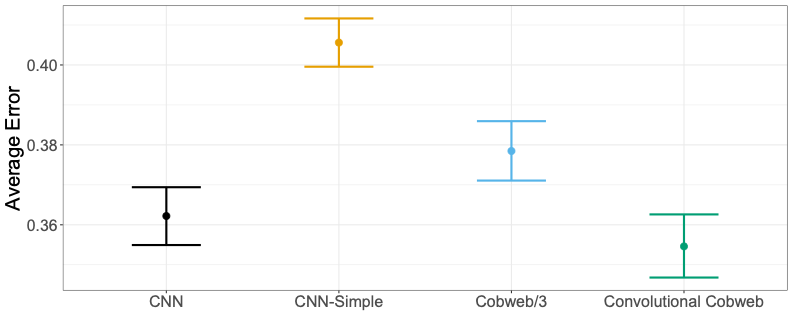}
    \caption{Overall error of the each approach on the incremental MNIST task, averaging over 50 runs. For each run, each system was presented with 300 randomly ordered images (30 images from each digit 0-9). The whiskers denote bootstrapped 95\% confidence intervals over runs.}
    \label{fig:overall_error}
\end{figure}

\begin{figure}
    \centering
    \includegraphics[width=\textwidth]{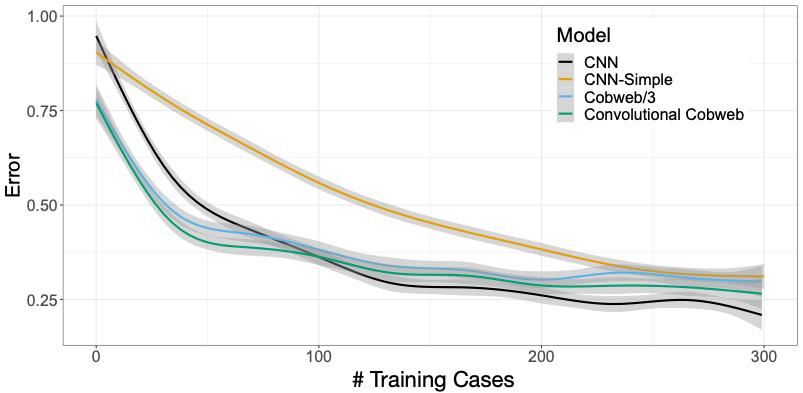}
    \caption{Learning curve showing the error of each approach at each point over the course of the incremental MNIST task. The lines were generated by applying Lowess smoothing to the 0-1 error data and the shaded regions denote 95\% confidence intervals. }
    \label{fig:learning_curves}
\end{figure}

\subsection{Results}
We applied each of our four approaches to the incremental MNIST digit recognition task. 
As mentioned previously, each approach was trained 50 times using a randomly sampled and shuffled sequence of 300 images (30 images from each digit). 
After training, we computed the average error rate over all of the predictions.
These error rates are shown in Figure \ref{fig:overall_error} along with 95\% confidence intervals estimated using bootstrapping \citep{wood2005bootstrapped}.\footnote{Bootstrapped confidence intervals provide an alternative to statistical significance tests that are generally viewed as more transparent.}
In addition to computing the overall error rates for each approach, we also wanted to understand how the performance changed over the course of incremental training.
Thus, we computed the learning curves for each approach, as presented in Figure \ref{fig:learning_curves}.
These curves show the average error rates at each point over the course of the training (e.g., the average error on the 100th training opportunity).

\subsection{Discussion}
Our overall results show that our new Convolutional Cobweb approach outperforms two of our baselines (see Figure \ref{fig:overall_error}). 
Specifically, we see that our approach outperforms Cobweb/3, which uses a similar incremental learning paradigm, but lacks a convolutional layer. 
We also see that it outperforms CNN-Simple, which has a similar convolutional structure, but differs in terms of its structure and learning paradigm.
When we compare Convolutional Cobweb to the more complex CNN, we see that there is not a meaningful difference in the overall error.

When we review the learning curve results (see Figure \ref{fig:learning_curves}), we find that, with the exception of CNN, these results largely hold over the entire duration of training.
We find that Convolutional Cobweb slightly outperforms Cobweb/3 over the entire course of training.
While it might not seem that the performance of Convolutional Cobweb is substantially better than Cobweb/3, models on the MNIST dataset generally tend to differ by fractions of a percent \citep{kowsari2018rmdl}.
Thus, this difference is a substantial improvement.
This claim is supported by the overall error results, where we can see that Convolutional Cobweb makes fewer errors than Cobweb/3.

Next, we look at the how our approach compares to the neural networks.
First, we find that our approach outperforms CNN-Simple over the entire range, suggesting it achieves better performance when compared to a CNN with a similar convolutional structure (one layer with one 3x3 convolutional filter).
However, we find that while Convolutional Cobweb outperforms the more complex CNN approach over the first 100 training cases, the CNN outperforms Convolutional Cobweb over the latter two thirds of training.
This suggests that the more complex CNN will likely outperform Convolutional Cobweb if we were to train it on a larger number of training cases.
This CNN may achieve such good performance because it has a substantially more complex convolutional structure (two layers with 2,080 3x3 convolutional filters). This suggests that future work should explore variants of Convolutional Cobweb that utilize more complex convolutional structures.

We know from our batch training evaluations of CNN-Simple and CNN that they can both achieve substantially better performance on the non-incremental MNIST digit recognition task (CNN-Simple has an error of 2.48\% and CNN has an error of 1.26\%) than we observe on the incremental MNIST task.
This suggests that there is room for substantial improvement on the incremental MNIST digit recognition task.
Based on our analysis, we believe the CNN approaches generally need more data and more stochastic gradient decent updates (i.e., more passes over the data) to achieve better performance.
In contrast, the Cobweb approaches tend to have faster learning rates and better performance on the earlier training opportunities, suggesting they do better when learning incrementally from fewer examples.

\section{Conclusion and Future Work}

Our findings show that Convolutional Cobweb demonstrates improved performance on the incremental MNIST digit recognition task over comparable baselines.
It outperforms one baseline model that uses a similar modeling paradigm, but that lacks convolutional capabilities (Cobweb/3).
It also outperforms a CNN model that has a similar convolutional structure (CNN-Simple).
Lastly, we find that while our approach outperforms a more complex CNN when training on fewer than 100 examples, the more complex CNN has better performance given a larger number of training cases. 
These findings suggest that integrating convolutional processing with incremental concept learning is a promising research direction; however, more work is needed to develop approaches at the intersection of these two lines of research.
In particular, future work should explore variants of Convolutional Cobweb that utilize multiple convolutional layers (similar to our more complex CNN model).
Our hope is that by utilizing more complex convolutional structures we can exceed the performance of the more complex CNN.

One major barrier to increasing the complexity of Convolutional Cobweb is its runtime performance.
Our CNNs were implemented using the PyTorch library, which lets them take advantage of matrix libraries and GPUs to greatly decrease the time it takes to process images and generate predictions.
For this work, we used one NVIDIA A40 GPUs to accelerate the CNN models.
These GPUs made it possible to do 150 epochs over 56,000 MNIST images in less than 7 minutes.
In contrast, Convolutional Cobweb does not utilize matrix libraries and does not support GPU acceleration.
As a result, it took approximately 5 hours to incrementally process 300 MNIST images on a 2GHz AMD EPYC 7662 CPU.
Increasing the convolutional complexity was computationally prohibitive.
Future work should explore how to implement the Cobweb algorithm in a library such as PyTorch, which would enable the use of both matrix libraries and GPU acceleration.

Additionally, the current work explores the application of Convolutional Cobweb to MNIST images, which only have a single input channel (images are black and white).
Future works should explore how to extend the approach to support multi-channel inputs, such as three channel RGB input.
There are multiple possible options for implementing multi-channel support that should be explored. For example, one extension might apply separate Convolutional Cobweb layers to each input channel, generating three output channels.
An alternative extension might apply a single convolutional layer to all three inputs and generate a single output channel.
Future work should explore these variants and evaluate them on multi-channel images, such as those found in the CIFAR-10 data set \citep{krizhevsky2009learning}.

In conclusion, our work explores the development of incremental approaches for learning from 2D visual images. We show that CNN models do not perform as well on incremental prediction tasks as they do on non-incremental prediction tasks. Additionally, we present Convolutional Cobweb, which merges the CNN concept of convolutions with the incremental learning approach used by Cobweb. This approach performs well on an incremental MNIST digit recognition task, outperforming a non-convolutional Cobweb approach and a CNN approach that uses a similar convolutional structure. This work represents a first step towards the unification of CNN advances with classic concept formation models.

 
\begin{acknowledgements} 
\noindent
We thank Pat Langley for his guidance, suggestions, and feedback while developing the Convolutional Cobweb model; this work would not have been possible without his input.
We also thank the Drexel University STAR program for providing support for Harshil Thakur to work on this project over the summer term.
Lastly, we thank the other members of the Teachable AI Lab (Qiao Zhang, Adit Gupta, and Glen Smith) for their suggestions and feedback on this work.
\end{acknowledgements} 

\vspace{-0.25in}

{\parindent -10pt\leftskip 10pt\noindent
\bibliographystyle{cogsysapa}
\bibliography{format}

\begin{thebibliography}{15}
\expandafter\ifx\csname natexlab\endcsname\relax\def\natexlab#1{#1}\fi
\expandafter\ifx\csname url\endcsname\relax
  \def\url#1{{\path{\sloppy #1}}}\fi
\expandafter\ifx\csname urlprefix\endcsname\relax\def\urlprefix{From }\fi

\bibitem[{Corter \& Gluck(1992)}]{corter1992explaining}
Corter, J.~E., \& Gluck, M.~A. (1992).
\newblock Explaining basic categories: Feature predictability and information.
\newblock {\em Psychological bulletin\/}, {\em 111\/}, 291.

\bibitem[{Deng(2012)}]{deng2012mnist}
Deng, L. (2012).
\newblock The mnist database of handwritten digit images for machine learning
  research.
\newblock {\em IEEE Signal Processing Magazine\/}, {\em 29\/}, 141--142.

\bibitem[{Fisher(1987)}]{fisher1987knowledge}
Fisher, D.~H. (1987).
\newblock Knowledge acquisition via incremental conceptual clustering.
\newblock {\em Machine learning\/}, {\em 2\/}, 139--172.

\bibitem[{Fisher et~al.(2014)Fisher, Pazzani, \& Langley}]{fisher2014concept}
Fisher, D.~H., Pazzani, M.~J., \& Langley, P. (2014).
\newblock {\em Concept formation: Knowledge and experience in unsupervised
  learning\/}.
\newblock Morgan Kaufmann.

\bibitem[{Gennari et~al.(1989)Gennari, Langley, \& Fisher}]{gennari1989models}
Gennari, J.~H., Langley, P., \& Fisher, D. (1989).
\newblock Models of incremental concept formation.
\newblock {\em Artificial intelligence\/}, {\em 40\/}, 11--61.

\bibitem[{Kingma \& Ba(2014)}]{kingma2014adam}
Kingma, D.~P., \& Ba, J. (2014).
\newblock Adam: A method for stochastic optimization.
\newblock {\em arXiv preprint arXiv:1412.6980\/}.

\bibitem[{Kowsari et~al.(2018)Kowsari, Heidarysafa, Brown, Meimandi, \&
  Barnes}]{kowsari2018rmdl}
Kowsari, K., Heidarysafa, M., Brown, D.~E., Meimandi, K.~J., \& Barnes, L.~E.
  (2018).
\newblock Rmdl: Random multimodel deep learning for classification.
\newblock {\em Proceedings of the 2nd International Conference on Information
  System and Data Mining\/} (pp. 19--28).

\bibitem[{Krizhevsky et~al.(2009)Krizhevsky, Hinton
  et~al.}]{krizhevsky2009learning}
Krizhevsky, A., Hinton, G., et~al. (2009).
\newblock {\em Learning multiple layers of features from tiny images\/}.
\newblock Technical report, University of Toronto.

\bibitem[{LeCun et~al.(2015)LeCun, Bengio, \& Hinton}]{lecun2015deep}
LeCun, Y., Bengio, Y., \& Hinton, G. (2015).
\newblock Deep learning.
\newblock {\em nature\/}, {\em 521\/}, 436--444.

\bibitem[{MacLellan et~al.(2016)MacLellan, Harpstead, Aleven, Koedinger
  et~al.}]{maclellan2016trestle}
MacLellan, C.~J., Harpstead, E., Aleven, V., Koedinger, K.~R., et~al. (2016).
\newblock Trestle: a model of concept formation in structured domains.
\newblock {\em Advances in Cognitive Systems\/}, {\em 4\/}, 131--150.

\bibitem[{McKusick \& Thompson(1990)}]{mckusick1990cobweb}
McKusick, K., \& Thompson, K. (1990).
\newblock {\em Cobweb/3: A portable implementation\/}.
\newblock Technical report, NASA Ames Research Center.

\bibitem[{McLure et~al.(2015)McLure, Friedman, \& Forbus}]{mclure2015extending}
McLure, M., Friedman, S., \& Forbus, K. (2015).
\newblock Extending analogical generalization with near-misses.
\newblock {\em Proceedings of the AAAI Conference on Artificial
  Intelligence\/}.

\bibitem[{Quinlan(1986)}]{quinlan1986induction}
Quinlan, J.~R. (1986).
\newblock Induction of decision trees.
\newblock {\em Machine learning\/}, {\em 1\/}, 81--106.

\bibitem[{Thompson \& Langley(1991)}]{thompson1991concept}
Thompson, K., \& Langley, P. (1991).
\newblock Concept formation in structured domains.
\newblock In {\em Concept formation\/},  127--161. Elsevier.

\bibitem[{Wood(2005)}]{wood2005bootstrapped}
Wood, M. (2005).
\newblock Bootstrapped confidence intervals as an approach to statistical
  inference.
\newblock {\em Organizational Research Methods\/}, {\em 8\/}, 454--470.

\end{thebibliography}

}


\end{document}